# Image Detection and Digit Recognition to solve Sudoku as a Constraint Satisfaction Problem

**Aditya Narayanaswamy, Yichuan Philip Ma, Piyush Shrivastava**

Khoury College of Computer Sciences

Northeastern University

Boston, MA 02115

{narayanaswamy.a, ma.yich, shrivastava.pi} @husky.neu.edu

**Abstract**

Sudoku is a puzzle well-known to the scientific community with simple rules of completion, which may require a complex line of reasoning. This paper addresses the problem of partitioning the Sudoku image into a 1-D array, recognizing digits from the array and representing it as a Constraint Satisfaction Problem (CSP). In this paper, we introduce new feature extraction techniques for recognizing digits, which are used with our benchmark classifiers in conjunction with the CSP algorithms to provide performance assessment. Experimental results show that application of CSP techniques can decrease the solution's search time by eliminating inconsistent values from the search space.

## 1. Introduction

Sudoku is a combinatorial puzzle in which numbers are be placed in a 9*9 grid, which is divided into 3*3 sub-grids. Some variations have a 12*12 grid with 4*3 sub-grids. The grid is partially completed, and each Sudoku has a distinct solution. The objective of the puzzle is to fill the grid with numbers 1-9, without the repetition of a number in a line, column or the sub-grid. Difficulty of the problem depends on the partially completed grid.

Constraint satisfaction can be defined as a process with a set of variables having constraints imposed on them that need to be satisfied. There are multiple real-life examples where CSP is utilized, such as automated planning & scheduling of classes for the semester. Representing Sudoku as a Constraint Satisfaction Problem and application of propagation techniques allow the puzzles to be solved with polynomial-time reasoning.

Fig. 1 Sudoku Puzzle

Retrieving text from images is a complex computer vision problem with several applications. One such application involves reading digits from a grid and parsing it into an array. Previous work on application of image processing techniques for a Sudoku image has demonstrated that that detection can be handled efficiently by a computer vision algorithm. Many feats have been accomplished in the field of pattern recognition, particularly with digit recognition due to availability of fast computers and learning algorithms. Efficiency of the machine learning algorithms have been rising but can be effectively boosted with the help of features. Simple features, like the morphological gradient where outline of the digits is highlighted, and the area in between is removed, can prove helpful, but certain other features like pixel count can help increase efficiency the most. Algorithms such as Boosted LeNet 4 in the past have been able to achieve efficiency greater than 99%, which is comparable to human performance.

Sudoku represented as a CSP in (Reeson, C. G., Huang, K. C., Bayer, K. M., & Choueiry, B. Y. 2007) proved to be an effective for study for our work. Our approach consists of image detection of 9*9 grid using OpenCV, digit recognition using machine and deep learning classifiers to solve the puzzle with CSP algorithms such as AC3 and AC4.

Our workflow is organized as follow: A discussion about related work has been presented in the next section, while in the next section there is a formal representation of the problem we're trying to solve. Section 3 describes the problem and discusses our approach to the problem. In the Section 4 experiments conducted with obtained results are presented, while in Section 5 we present the datasets used. We conclude the work in Section 6.

## 2. Related Work

An interesting paper that mentions recognition of numbers from the Sudoku image is (Simha, Pramod J. et al. 2012) which provides an algorithm to process the Sudoku image from any digital camera and form a virtual grid. (LeCun, et al. 1990) introduce a convolutional network for recognizing handwritten digits, while (Schölkopf, Bernhard 2002) demonstrated the use of kernels with support vector machines for optimization. (Liaw, Andy 2002) show the classification of digits using random forest.

Approaches such as (Simonis 2005) and (Lambert, T. et al. 2006) present Sudoku as a Constraint satisfaction Problem, with the application of distinct search methods to solve the puzzle. Demonstration of problem formulation and application of search and inference has been shown in (O'Sullivan, B., & Horan, J. 2007). (Bessiere, Christian 1994) present the implementation of CSP using arc-consistency.

Other interesting works are (O.D. Trier et al. 1996), (Dash, Manoranjan et al. 1997) and (Soto, Ricardo, et al. 2013). The first and second presented feature extraction methods for character recognition, while the third one a hybrid AC3 algorithm. (LeCun, Y. et al. 1995) presents a comparison of classification techniques for digit recognition.

## 3. Problem Description

In order to solve the sudoku problem, we first have to split the problem up into 3 parts, the computer vision part where the image is extracted and the sudoku numbers have been segregated, the identification part where the machine learning is implemented and the digits have been identified as their respective number, and the solving part where the CSP is used based on existing numbers and the output is produced in a 9*9 fully solved sudoku puzzle.

### 3.1 Computer Vision

Sudoku is first given to the system as an image which is then extracted by the computer vision algorithms. We take the image and greyscale it to process the squares. With the contours taken, the individual squares are processed. The contours are estimated by using Topological Structural Analysis of Digitized Binary Images by Border Following. The squares are then isolated, and the features are also extracted. The extracted features here are the morphological gradient and the Pixel Count based features.

Morphological transformations are operations based on the image shape. They are usually performed on binary images. It needs two inputs, one is our original image, second one is called structuring element or kernel which decides the nature of operation. Two basic morphological operators are Erosion and Dilation. Morphological gradient is the difference between erosion and dilation.

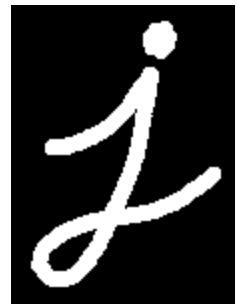
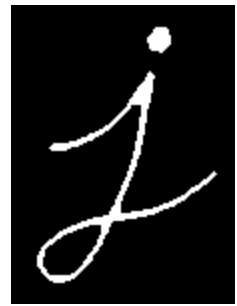

Fig. 2a Original Image       Fig. 2b Erosion

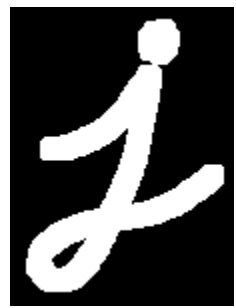
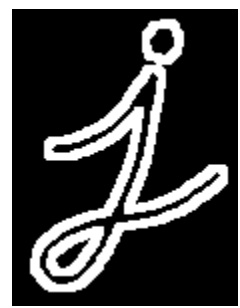

Fig. 2c Dilation              Fig. 2d Gradient

This gradient helps to take in only the image contours and process them for better accuracy.

Another feature used is the Pixel Count which is obtained by counting row-wise the number of black pixels present and doing the same column-wise, thus obtaining two profiles. For example, the row profile r of dimensions (1xN) can be obtained from complemented binary image l of NxN pixels, where 0s and 1s represent white and black pixels, respectively, by the following operation:

$$ri = \sum_{j=i}^{N} I(j,i) \quad i = 1,\ldots N \quad (1)$$

## 3.2 Identification

This section involves digit identification through machine learning and the deep learning models that have been used to place the digits in 81 separate squares.

### 3.2.1 K-Nearest Neighbor

The approach was proposed by (T.M. Cover and P.E. Hart in 1967). K-NN algorithm uses the distance equations to determine if the input belongs to a classification. During the training phase, it does not compute anything, but just stores all the training data information for reference during the fitting phase. During that phase, it calculates the distance between every training data input parameter with the given input data and tries to isolate the ones with the least distance between the input. After this, it takes the first N closest training data and finds classification which is the most frequent amongst the N closest training data. The function for the distance can be set to any appropriate functions like the Manhattan distance. In the experiments, the Euclidean distance was used for the comparison.

$$D(p,q) = \sqrt{\sum_{i=1}^{n}(qi - pi)^2} \quad (2)$$

### 3.2.2 Support Vector Machine

The SVM classifier plots all the given training data and tries to calculate an imaginary hyperplane that separates all the different classified training data. It uses this imaginary plane to segregate the incoming values in the testing phase and classifies the input using this method. The classification of the input data is based on which side of this hyperplane the data falls into. Certain classifications can be segregated using basic linear planes whereas other classifications are not linear separated and in which case, they will be extrapolated onto a new dimension and separated in the higher order dimension and then brought back to the current plane. These can be done using kernels. The kernels used here are the linear (3), polynomial (4) and Radial Basis Function (RBF) (5).

$$K(x, x') = \exp(-\frac{\|x-x'\|^2}{2\sigma^2}) \quad (3)$$

$$K(\vec{x}, \vec{x'}) = \vec{u} \cdot \vec{v} \quad (4)$$

$$K(\vec{x}, \vec{x'}) = (\vec{u} \cdot \vec{v} + b)^n \quad n > 1 \quad (5)$$

### 3.2.3 Convolutional Neural Network

CNNs are an extension of multilayer perceptrons, which can learn filters that need to be computed by the machine learning models, as experimented earlier in [Yann LeCun et al. 1989] using back-propagation. Convolutional networks are mainly applied on visual imagery. Since the training process involves learning about patterns from smaller patterns, computations usually are time-consuming and may require GPU-based implementation. CNNs work well with small-size images, but with large-resolution images, weights with full-scale connectivity cannot be processed efficiently.

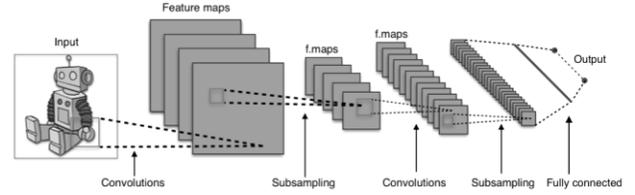

Fig 3. CNN Architecture

A convolutional neural network consists of distinct hidden layers in addition to the input and output layers. These distinct layers usually consist of convolutional layer with filters that can be learned, rectified linear unit layer for application of activation function, pooling layer for down-sampling and loss layer for specification of penalization for incorrect output. We use Keras with TensorFlow as backend to process the CNN model and provide a comparison between the distinct CNN implementations formed by choosing different hyperparameters associated with each layer.

#### 3.2.3.1 Hyperparameter Tuning

Selection of hyperparameters often characterizes the performance of a CNN. Cross-validation cannot be performed due to the long training time, so tuning of following parameters was performed using the GridSearch method for a CNN with different layers:

- Batch size and epochs – The batch size defines the number of patterns to be read and stored in the memory. Evaluation was done in batch size from 32 to 128 in steps of 32. Epochs were tuned from 10 to 60 in steps of 10.
- Optimizer – Optimizers such as Stochastic Gradient Descent, Adam and Adamax were passed while constructing the CNN model.
- Learning rate – Learning rate allows the weights to be updated at pre-defined rate. Momentum is another component which controls the influence of previous weights. Learning rate was va ried from 0.001 to 0.5, while for momentum we tried using values from 0.2 to 1.
- Activation function – The activation function decides when neurons will be fired. Different action functions such as softmax, relu and tanh were assessed for tuning.
- Weights – Different weight initialization schemes such as normal and glorot_normal were used for evaluating the performance through tuning.

### 3.3 Solving the puzzle

To solve the sudoku puzzles, we represented them as Constraint Satisfaction Problems (CSPs). In a typical game of sudoku, with a 9*9 grid for a puzzle, the goal is to fill in all the squares with digits 1-9 so that each row has unique values assigned, along with each column and each 3x3 sub-grid. The variables for the puzzles were the squares that were initially empty (to be filled out as the game progresses), and the domain for each of those variables were the individual decimal digits 1-9. Furthermore, the constraints for Sudoku were 9-way alldiff constraints, where each value is unique to each variable, for each row (for a total of 9 constraints), each column (9 more constraints), and each 3*3 sub-grid (9 more constraints), for a total of 27 alldiff constraints. We used the backtracking search algorithm with enhancements, which we discuss next, to solve the sudoku CSPs.

#### 3.3.1 Backtracking Search

To solve the sudoku CSPs, we used the Backtracking Search algorithm. This algorithm is a version of depth-first search (DFS) with the modifications of checking and assigning a value to one variable at a time and of checking constraints after each variable assignment. It is the standard algorithm used to solve any CSP (Kumar 1992), not just Sudoku. The following improvements to Backtracking that we chose to use, maintaining arc consistency (MAC) and heuristics for selecting the next variable for assignment, which we will discuss next, greatly improve the performance of Backtracking in searching for a solution to a CSP.

#### 3.3.2 Minimum Remaining Values (MRV) Heuristic

Heuristics for expanding the next variable are typically used in backtracking search to improve its performance and to detect failures sooner. The Minimum Remaining Values (MRV) heuristic is used to choose and expand the variable with the least amount of available values in its domain when choosing the next variable for assignment (Kumar 1992). There are other heuristics, such as least constraining value (LCV) and the degree heuristic; however, in this paper, we chose to use the MRV heuristic to pick the next square in a Sudoku puzzle given how Sudoku is typically played.

#### 3.3.3 Arc Consistency

To further improve backtracking, we used the strategy of maintaining arc consistency (MAC). For any arc between two variables, x and y, in a CSP, if there is a value in the domain of y that satisfies the constraint between the two variables for any value in the domain of x. Generally, if all arcs in a CSP are consistent, then the CSP is arc-consistent. In our implementation of backtracking search, we enforced arc consistency after each time a variable was assigned a value.

There are several algorithms that enforce arc consistency on a CSP. They have different time and space complexities despite that they were invented to do the same task (Mackworth 1977). The ones that we have implemented and experimented with are AC-1 (Mackworth 1977), AC-2 (Mackworth 1977), and AC-4 (Mohr and Henderson 1986). We have also implemented AC-3, the most commonly used arc consistency algorithm, but we only used that as a control to experiment with our other arc consistency algorithm implementations. AC-1, AC-2, and AC-3 use the following Revise helper method (Mackworth 1977), whose pseudocode is as follows:

Procedure Revise($X_i$, $X_j$) returns a Boolean value:
1. revised = false
2. for each $a_i$ in domain[$X_i$]:
3.    if there is no $a_j$ in domain[$X_j$] that satisfies a constraint:
4.       remove $a_i$ from domain[$X_i$]
5.       revised = true
6. return revised

The pseudocode for AC-1 is as follows.

Procedure AC-1(CSP) returns a Boolean value:
1. repeat:
2.    for ($x_i$, $x_j$) in CSP's arcs:
3.       Revise($x_i$, $x_j$) and Revise($x_j$, $x_i$)
4.       if no values remain in either domain of $x_i$ or $x_j$, return false
5. until no domain has changed
6. return true

The pseudocode for our implementation of AC-2 is as follows.

Procedure AC-2(CSP) returns a Boolean value:
1. for i=0 until 81: // for the total squares in a sudoku puzzle
2.    Q1 = empty queue
3.    Q2 = empty queue
4.    for j=0 until i:
5.       push (CSP.variables[i], CSP.variables[j]) into Q1
6.       push (CSP.variables[j], CSP.variables[i]) into Q2
7.    while Q1 is not empty:
8.       while Q1 is not empty:
9.          pop ($x_i$, $x_j$) from Q1
10.          if Revise($x_i$, $x_j$):
11.             if the domain of $x_i$ has no remaining values, return false
12.             for j=0 until i:
13.                if $x_j \neq$ CSP.variables[j], push (CSP.variables[j], $x_i$) into Q2
14.    Q1 = copy of Q2
15.    clear Q2

16. return true

Because AC-3 is very commonly used as the algorithm for enforcing arc consistency in CSPs, we have decided to omit its pseudocode. The pseudocode for our implementation of AC-4 is as follows.

Procedure AC-4(CSP) returns a Boolean value:
1. Q = empty queue
2. supports = empty map // set of key-value pairs, as in a Python dictionary
3. counter = empty map
4. for $(x_i, x_j)$ in CSP.arcs, $a_i$ in CSP.domains[$x_i$], $a_j$ in CSP.domains[$x_j$]:
5.     supports[$(x_i, a_i)$] = empty list
6.     counter[$(x_i, a_i, x_j)$] = 0
7. for $(x_i, x_j)$ in CSP.arcs and $a_i$ in CSP.domains[$x_i$]:
8.     for $a_j$ in CSP.domains[$x_j$]:
9.         if $a_i$ and $a_j$ satisfy a constraint:
10.            increment counter[$(x_i, a_i, x_j)$]
11.            Add $(x_i, a_i)$ to supports[$(x_j, a_j)$]
12.     if counter[$(x_i, a_i, x_j)$]=0:
13.        push $(x_i, a_i)$ to Q
14.        remove $a_i$ from CSP.domains[$x_i$]
15.     if CSP.domains[$x_i$] is empty, return false
16. while Q is not empty:
17.    pop $(x_j, a_j)$ from Q
18.    for $(x_i, a_i)$ in supports[$(x_j, a_j)$]:
19.        if CSP.domains[$x_i$] has $a_i$:
20.            decrement counter[$(x_i, a_i, x_j)$]
21.            if counter[$(x_i, a_i, x_j)$]=0:
22.                push $(x_i, a_i)$ to Q
23.                remove $a_i$ from CSP.domains[$x_i$]
24.        if CSP.domains[$x_i$] is empty, return false
25. return true

There are more arc consistency algorithms, such as AC-6 (Bessière 1994), but we will save them for future work.

## 4. Datasets

Publicly available datasets from UCI repository, MNIST repository and Kaggle were used to retrieve the digits dataset, consisting of images and the resultant digit they have, to recognize the digits in the picture. The Chars74 dataset has over 7000 images of printed characters with dimension 128*128 for different digits. The dataset proved effective as we were able to obtain greater accuracy than achieved with the MNIST dataset with our features. Sudoku images for the experiments were chosen from the public domain of commons repository.

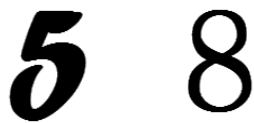

Fig. 4 Sample images from Char74 dataset

## 5. Experiements

**5.1 Machine Learning and Digit Recognition**

Trained K-NN and SVC models were tested first on four simple sudoku squares. K-NN models were trained with the Euclidean distance with the neighbor settings at 5, 10, 20, 50 and 100 with the basic input wherein there was a white background and the digits were black, an inverted version of the basic image aimed at removing background noise, morphological gradient and the pixel count features. The results were as below:

Table 1. K-NN model results.

| K-NN with feature | Accuracy [In %] |
|---|---|
| 5N with Basic | 92.59 |
| 5N with Inverted | 92.59 |
| 5N with Morphological | 93.21 |
| 5N with PCF | 87.96 |
| 10N with basic | 91.67 |
| 10N with inverted | 91.67 |
| 10N with morphological | 93.21 |
| 10N with PCF | 87.65 |
| 20N with basic | 91.35 |
| 20N with inverted | 91.35 |
| 20N with morphological | 92.90 |
| 20N with PCF | 88.27 |
| 50N with basic | 91.35 |
| 50N with inverted | 91.36 |
| 50N with morphological | 93.21 |
| 50N with PCF | 87.34 |
| 100N with basic | 92.28 |
| 100N with inverted | 92.28 |
| 100N with morphological | 92.59 |

| 100N with PCF | 87.03 |

With the SVC model, training was done with the same features with three different kernels - Linear, Poly and Radial Basis Function. The results were as below.

Table 2. SVC Model Results.

| Kernel with Feature | Accuracy [In %] |
|---|---|
| Linear with basic | 93.21 |
| Linear with inverted | 93.21 |
| Linear with morphological | 92.29 |
| Linear with PCF | 85.80 |
| Poly with basic | 93.21 |
| Poly with inverted | 93.21 |
| Poly with morphological | 85.80 |
| Poly with PCF | 92.29 |
| RBF with basic | 92.28 |
| RBF with inverted | 92.28 |
| RBF with morphological | 93.21 |
| RBF with PCF | 89.81 |

Pixel Count Feature seems to be performing lower than the others as it might be due to overfitting and for a relatively easy input with little complication, it may not be the choice. Morphological feature performs well in contrast as it removes the extra pixels covering the digit and only takes the contours, thereby reducing the chances of mistaking the digit with another similar for example 1 and 7.

**5.2 Deep Learning and Digit Recognition**
CNN models with different convolutional and pooling layers were first tested on the sudoku squares with relu as the activation function, batch size of 64 and 20 epochs after transforming the data. The results were as follows:

Table 3. Accuracy for CNN with different layers

| CNN with layers | Accuracy [In %] |
|---|---|
| 1 Conv2D layer | 98.35 |
| 2 Conv2D layers | 99.48 |
| 3 Conv2D layers | 99.61 |
| 4 Conv2D layers | 98.96 |

Initially the accuracy increases after adding $2^{nd}$ and $3^{rd}$ layer, but it seems to drop after adding the $4^{th}$, which is expected due to overfitting.

Tuning was performed with the CNN model for parameters belonging to different layers with distinct functions. Mean score from the results was chosen as the comparison metric. The results are as illustrated:

Table 4. Mean Score for Batch size and epoch tuning

| Batch size | Epochs | Mean Score |
|---|---|---|
| 32 | 20 | 98.73 |
| 32 | 40 | 98.92 |
| 32 | 60 | 99.03 |
| 64 | 20 | 99.61 |
| 64 | 40 | 99.67 |
| 64 | 60 | 99.78 |
| 96 | 20 | 98.91 |
| 96 | 40 | 99.07 |
| 96 | 60 | 99.16 |
| 128 | 20 | 99.29 |
| 128 | 40 | 99.35 |
| 128 | 60 | 99.57 |

The tuning illustrates that batch size of 64 achieves the best results with 60 epochs. Testing the CNN with different optimization algorithms gave us the results below.

Table 5. Mean Score for Optimization algorithm tuning

| Optimizer | Mean Score |
|---|---|
| SGD | 0.81 |
| Adamax | 0.92 |
| Adagrad | 0.95 |
| RMSprop | 0.93 |
| Adadelta | 0.94 |
| Adam | 0.93 |
| Nadam | 0.95 |

'Nadam' yields the best performance and can be considered as the most efficient optimizer for our data. Next, activation function was tuned with the default CNN model.

Table 6. Mean score for tuning of activation function

| Activation Function | Mean Score |
| --- | --- |
| softsign | 0.943 |
| relu | 0.945 |
| tanh | 0.947 |
| sigmoid | 0.942 |
| softmax | 0.949 |
| softplus | 0.943 |
| hard_sigmoid | 0.940 |
| linear | 0.943 |

The function 'softmax' performs the best here, while other activation functions are a close approximate.

Table 7. Mean score for tuning of learning rate and momentum

| Learning Rate | Momentum | Mean Score |
| --- | --- | --- |
| 0.001 | 0.2 | 0.161 |
| 0.001 | 0.4 | 0.225 |
| 0.001 | 0.6 | 0.354 |
| 0.001 | 0.8 | 0.561 |
| 0.001 | 0.9 | 0.810 |
| 0.001 | 1 | 0.902 |
| 0.01 | 0.2 | 0.853 |
| 0.01 | 0.4 | 0.902 |
| 0.01 | 0.6 | 0.925 |
| 0.01 | 0.8 | 0.934 |
| 0.01 | 0.9 | 0.951 |
| 0.01 | 1 | 0.362 |
| 0.1 | 0.2 | 0.140 |
| 0.1 | 0.4 | 0.117 |
| 0.1 | 0.6 | 0.112 |
| 0.1 | 0.8 | 0.126 |
| 0.1 | 0.9 | 0.111 |
| 0.1 | 1 | 0.101 |
| 0.2 | 0.2 | 0.118 |
| 0.2 | 0.4 | 0.104 |
| 0.2 | 0.6 | 0.111 |
| 0.2 | 0.8 | 0.280 |
| 0.2 | 0.9 | 0.123 |
| 0.2 | 1 | 0.114 |
| 0.3 | 0.2 | 0.112 |
| 0.3 | 0.4 | 0.140 |
| 0.3 | 0.6 | 0.121 |
| 0.3 | 0.8 | 0.101 |
| 0.3 | 0.9 | 0.185 |
| 0.3 | 1 | 0.109 |
| 0.4 | 0.2 | 0.102 |
| 0.4 | 0.4 | 0.113 |
| 0.4 | 0.6 | 0.092 |
| 0.4 | 0.8 | 0.121 |
| 0.4 | 0.9 | 0.144 |
| 0.4 | 1 | 0.132 |
| 0.5 | 0.2 | 0.107 |
| 0.5 | 0.4 | 0.114 |
| 0.5 | 0.6 | 0.110 |
| 0.5 | 0.8 | 0.118 |
| 0.5 | 0.9 | 0.140 |
| 0.5 | 1 | 0.113 |

Learning rate 0.01 and the momentum 0.9 achieve the best results for digit recognition. As expected, on increasing the learning rate beyond 0.1, momentum does not have significant impact on the mean score as the rate tends to stay close to 0.11.

Table 8. Mean score for tuning of weight initialization

| Wight Initialization Scheme | Mean Score |
| --- | --- |
| uniform | 0.949 |
| normal | 0.944 |
| zero | 0.853 |
| glorot_normal | 0.941 |
| glorot_uniform | 0.945 |
| he_normal | 0.945 |
| he_uniform | 0.945 |
| lecun_uniform | 0.946 |

Inference from the results indicates that the 'uniform' weight initialization scheme achieves the highest mean score.

### 4.2 Arc Consistency Algorithms Comparison

We implemented arc consistency algorithms AC-1, AC-2, and AC-4 according to the pseudocode in section 3.3.3. To experiment with them, we also implemented AC-3 to use as a control. We used five sample sudoku grids, each one varying in difficulty to test our CSP implementation with each of the arc consistency algorithms. For each of the grids we had, we ran each algorithm on it three times, recording the runtime for each run. The results from the runs, which are as below, were the average runtimes of the three trials for each algorithm on each grid.

Table 9: Arc Consistency Algorithm Results (runtimes in seconds)

| Arc Consistency Algorithm | Sudoku Grid | | | | |
|---|---|---|---|---|---|
| | Sudoku1 | Sudoku2 | Sudoku3 | Sudoku4 | Sudoku5 |
| AC-3 (control) | 10.46 | 172.29 | 0.43 | 34.63 | 1.73 |
| AC-1 | 10.74 | 179.87 | 0.43 | 32.36 | 1.67 |
| AC-2 | No Solution Found | | | | |
| AC-4 | 23.09 | 316.07 | 0.79 | 57.63 | 3.52 |

The more difficult a puzzle is, the longer it takes to solve it. With these results, that statement is true not just for humans, but for computers as well. One can clearly see from the results that Sudoku2 was a notoriously difficult puzzle to solve, whereas Sudoku3 was a very easy one.

From our results, AC-1 produced similar runtimes as those of AC-3 to solve each of the sudoku puzzles, so we can conclude that for common 9*9 sudoku grids, AC-1 for arc consistency is about as effective as AC-3, except that AC-1 uses less space than AC-3, since AC-3 uses a FIFO queue to keep track of all arcs in a CSP whereas AC-1 does not (Mackworth 1977). AC-1, however, may not be as efficient in solving other kinds of Constraint Satisfaction Problems. Yet the simplicity of AC-1 has proved to be effective in solving sudoku.

Because running backtracking with AC-2 on all the sudoku puzzles that we had failed to solve any of them, we can conclude that AC-2 is not suitable for solving sudoku at all. This is because AC-2 enforces arc consistency on all arcs in the puzzle, not just those that connect two neighboring variables, such as two squares that are in totally different rows, columns, and 3*3 sub-grids, which are obviously not neighbors of each other. With AC-2, a certain variable's domain is emptied more quickly and unnecessarily than with AC-1 or AC-3, thus leading to AC-2 failing to solve any sudoku puzzle.

Also, from our results, AC-4 took about twice as much time as AC-3 to solve our sample puzzles, so we can conclude that AC-4 is inefficient for solving Sudoku. Aside from the extra data structures of the support map and the counter, leading to using superfluous amounts of space, we attribute the inefficiency of AC-4 to it being complicated to understand and rather tricky to implement. Despite the claims of the optimality of AC-4 (Mohr and Henderson 1986), it does not seem so when it comes to solving Sudoku.

### 5.Conclusion

In this work we have proposed methods for identification of a Sudoku image, recognition of digits from the partial grid and solving the puzzle by representing as a Constraint Satisfaction Problem. We have observed that the combined use of these algorithms have highly useful applications.

We were able to demonstrate through our experiments that CNNs provide higher accuracy for digit recognition, while SVM and KNN models depends on custom features for achieving comparable results. Future work may involve adding more features for the machine learning models and adding classifiers to have broader benchmark system.

For solving Sudoku CSPs using backtracking search with maintaining arc consistency and the MRV heuristic, we can conclude for now that AC-1 is the best algorithm for enforcing arc consistency in Sudoku CSPs due to its similar runtimes with AC-3 and lower space utilization as compared to AC-3. We will try to implement and experiment with other arc consistency algorithms, such as AC-6 and approaches involving simulated annealing.